\documentclass[pdflatex,sn-mathphys]{sn-jnl}

\usepackage{graphicx}
\usepackage{multirow}
\usepackage{mathtools}
\usepackage{tabularx}
\usepackage{hyperref,xcolor}
\usepackage[flushleft]{threeparttable}

\jyear{2021}%

\theoremstyle{thmstyleone}%
\theoremstyle{thmstyletwo}%

\theoremstyle{thmstylethree}%

\begin{document}

\title[Calibration Error Estimation Using Fuzzy Binning
]{Calibration Error Estimation Using Fuzzy Binning
}


\author{\fnm{Geetanjali} \sur{Bihani}}\email{gbihani@purdue.edu}

\author{\fnm{Julia} \sur{Taylor Rayz}}\email{jtaylor1@purdue.edu}

\affil{\orgdiv{Computer and Information Technology}, \orgname{Purdue University}, \orgaddress{\street{}, \city{West Lafayette}, \postcode{47904}, \state{Indiana}, \country{USA}}}


\abstract{Neural network-based decisions tend to be overconfident, where their raw outcome probabilities do not align with the true decision probabilities. Calibration of neural networks is an essential step towards more reliable deep learning frameworks. Prior metrics of calibration error primarily utilize crisp bin membership-based measures. This exacerbates skew in model probabilities and portrays an incomplete picture of calibration error. In this work, we propose a Fuzzy Calibration Error metric (FCE) that utilizes a fuzzy binning approach to calculate calibration error. This approach alleviates the impact of probability skew and provides a tighter estimate while measuring calibration error. We compare our metric with ECE across different data populations and class memberships. Our results show that FCE offers better calibration error estimation, especially in multi-class settings, alleviating the effects of skew in model confidence scores on calibration error estimation. We make our code and supplementary materials available at: \href{https://github.com/bihani-g/fce}{https://github.com/bihani-g/fce}}.


\maketitle

\section{Introduction}\label{sec1}
Neural network-based decision-making systems have evolved rapidly in the recent decade. Within the domain of natural language processing, deep learning has shaped the current evolution in language modeling. These neural network-based language models are trained on large text corpora and can be fine-tuned across a wide range of NLP tasks and further improved using synthetic semantic enhancement schemes \cite{bihani2021low}, yielding state-of-the-art performance \cite{chen2019bert, devlin2018bert, radford2019language, yang2019xlnet}. Ideally, a neural model should output reliable and confident prediction probabilities. But recent works have shown that neural networks are unreliable and output highly overconfident predictions, resulting in over-estimation of the model's confidence in decisions \cite{guo2017calibration, kong_calibrated_2020, jiang_how_2021}. This leads to model miscalibration, i.e. a lack of alignment between a model’s decision probabilities and its actual likelihood of correctness. This lack of calibration can severely impact the trustworthiness of a model's decisions. 

A widely adopted measure of the degree of miscalibration is Expected Calibration Error (ECE) \cite{naeini2015obtaining}, used to measure neural network reliability \cite{ovadia2019can, huang2020tutorial, tack2020csi}. The highly overconfident output prediction probabilities of neural networks result in a left-skewed probability distribution \cite{nixon_measuring_nodate}. Since ECE utilizes a fixed-width crisp binning scheme, this skew results in higher probability bins largely contributing to the calibration error estimation, while lower probability bins are ignored \cite{nixon_measuring_nodate, roelofs2022mitigating, ding2020revisiting}. To overcome these limitations, prior works have proposed alternative binning strategies such as equal-frequency binning \cite{roelofs2022mitigating}, adaptive binning \cite{ding2020revisiting}, replacing binning with smoothed kernel density estimation \cite{zhang2020mix}, and more. Most calibration error estimation techniques rely on crisp binning, which discards edge probabilities (probabilities that typically lie on the bin edge) that could have contributed to a more accurate calibration error estimation. Although some works have utilized fuzzification of prediction probabilities for downstream NLP tasks \cite{bihani2022fuzzy}, the calibration impacts of such fuzzification are yet to be studied. We hypothesize that fuzzifying the binning scheme would allow edge probabilities to contribute toward more accurate calibration error estimation. Moreover, fuzzy binning would increase the visibility of lower probability scores by allowing them to have partial membership in higher probability bins, minimizing the skew problem in calibration error estimation.

Towards testing this hypothesis, we propose a new metric for estimating calibration error, i.e. Fuzzy Calibration Error (FCE), that utilizes fuzzy binning instead of crisp binning to allow edge probability contributions and minimize skew in calculating calibration error. We perform empirical evaluation across different classification settings, comparing FCE with the baseline calibration error estimation metric ECE. 

Our results show that, unlike ECE, FCE better captures miscalibration in lower probability bins and provides a tighter and less skewed estimate of calibration error. These improvements are more visible in multi-class settings, where the skew in confidence scores exacerbates the calibration error estimation problem. 

The contributions of this work are summarized as follows: 
\begin{itemize}
    \item We propose Fuzzy Calibration Error (FCE) metric which uses fuzzy binning to account for edge probabilities and minimize skew in calibration error estimation 
    \item We perform empirical evaluation across a wide range of classification settings and show the benefits of using FCE over ECE in minimizing the impacts of probability skew on calibration error estimation
\end{itemize}

\section{Background}\label{sec2}
\subsection{Neural Network Calibration}
Neural network calibration refers to the process of adjusting a neural network model's output probabilities to reflect the true probabilities of the events it is predicting. With the increased application of neural network architectures in high-risk real-world settings. their calibration has become an extensively studied topic in recent years \cite{thulasidasan2019mixup, malinin2018predictive, hendrycks_pretrained_2020}. Recent research has focused on improving the calibration of neural networks, particularly in the context of deep learning. Various methods have been proposed to achieve better calibration, including temperature scaling \cite{guo2017calibration}, isotonic regression \cite{platt1999probabilistic}, and histogram binning \cite{zadrozny2001obtaining}.

\subsection{Expected Calibration Error}
Expected calibration error (ECE) is a scalar measure of calibration error that calculates the weighted average of the difference between the accuracy of a model and its average confidence level over a set of bins defined by the predicted probabilities. Estimation of expected accuracy from finite samples is done by grouping predictions into $M$ interval bins (each of size $\frac{1}{M}$), and the accuracy of each bin is calculated. Let $B_{m}$ be a bin containing samples whose prediction confidence lies within the interval $I_{m}=\left(\frac{m-1}{M}, \frac{m}{M}\right]$. Then the accuracy of $B_{m}$, where $y_{i}$ and $\hat{y}_{i}$ portray predicted and true class labels, is calculated as shown in Eq. \ref{acc_cal}.

\begin{equation}\label{acc_cal}
\operatorname{acc}\left(B_{m}\right)=\frac{1}{\left\vert B_{m} \right\vert}\sum_{i \in B_{m}} \mathbf{1}\left(\hat{y}_{i}=y_{i}\right)
\end{equation}

The average predicted confidence of $B_{m}$, is calculated as shown in Eq. \ref{conf_cal}, where $\hat{p}_{i}$ refers to the prediction probability of the $i^{th}$ instance in $B_{m}$.

\begin{equation}\label{conf_cal}
\operatorname{conf}\left(B_{m}\right)=\frac{1}{\left\vert B_{m} \right\vert}\sum_{i \in B_{m}} \hat{p}_{i}
\end{equation}

In an ideal scenario, for a perfectly calibrated model, $\operatorname{acc}\left(B_{m}\right) = \operatorname{conf}\left(B_{m}\right)$ for all $m$ bins where $m \in\{1, \ldots, M\}$.

Finally, ECE is calculated as shown in Eq. \ref{ece}, where $n$ is total number of samples \cite{naeini2015obtaining}.

\begin{equation}\label{ece}
\mathrm{ECE}=\sum_{m=1}^{M} \frac{\left\vert B_{m}\right\vert}{n} \mid \operatorname{acc}\left(B_{m}\right)-\operatorname{conf}\left(B_{m}\right)
\end{equation}

\section{Fuzzy Calibration Error}\label{sec3}
In this work, we propose Fuzzy Calibration Error (FCE), a metric that transforms raw prediction probabilities into soft bin membership values for calibration error estimation. This transformation has two benefits:
\begin{enumerate}
    \item Allows edge probability contributions when calculating calibration error
    \item Minimize probability skew effects by increasing visibility of lower probability bins in calibration error estimation
\end{enumerate}

To perform fuzzification, we utilize trapezoidal membership functions to map raw softmax prediction probabilities to fuzzy bin membership values. The difference between crisp and fuzzy binning of model prediction probabilities is shown in Figure~\ref{fuzzyvscrips}, with $M=3$ bins, and can be extended to any number of bins where $M>3$. While ECE only allows for crisp membership within each bin, FCE offers a more flexible binning approach, with partial memberships allowed across multiple bins. 
\begin{figure}
    \centering
    \begin{minipage}[b]{0.3\textwidth}
        \includegraphics[width=0.9\textwidth]{figures/crisp.pdf}
    \vfill
        \includegraphics[width=0.9\textwidth]{figures/fuzzy.pdf}
    \end{minipage}
    \includegraphics[width=0.5\textwidth]{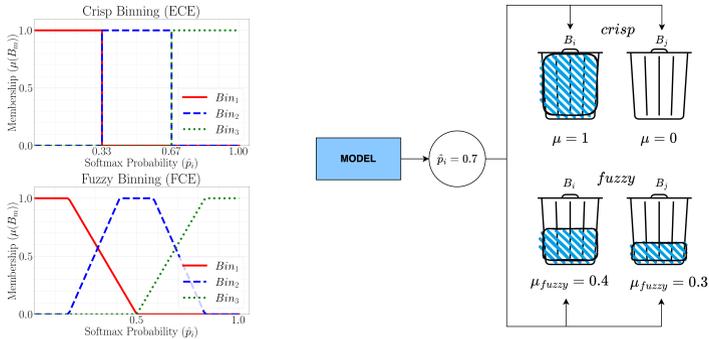}
    \caption{Crisp binning (Top left) and fuzzy binning (Bottom left) of prediction probabilities, where the number of bins $M=3$. An example of the difference in bin assignment based on $\hat{p_{i}}$ in crisp vs fuzzy binning (Right).}
\end{figure}\label{fuzzyvscrips}

Fuzzy Calibration Error ($FCE$) calculates the weighted average of the difference between accuracy and average model confidence over a set of $M$ fuzzy bins. Estimation of expected accuracy from finite samples is done by grouping predictions into $M$ fuzzy bins, and the accuracy of each bin is calculated. Let $B_{m}$ be a bin containing samples whose prediction confidence lies within the interval $I_{m}=\left(\frac{m-1}{M}, \frac{m}{M}\right]$. Then the accuracy for bin $B_{m}$, where $y_{i}$ and $\hat{y}_{i}$ portray predicted and true class labels, is calculated as shown in Eq. \ref{fuzzy_acc}.

\begin{equation}\label{fuzzy_acc}
\operatorname{acc}{_{fuzzy}}(B_{m})=\frac{1}{\lvert{\mu_{fuzzy}(B_{m})}\rvert} \sum_{i \in B_{m}} \mu_{fuzzy}(B_{m})(\hat{y}_{i}=y_{i})
\end{equation}

Then, the average fuzzy predicted confidence of $B_{m}$, is calculated as shown in Eq. \ref{fuzzy_conf}.

\begin{equation}\label{fuzzy_conf}
\operatorname{conf}{_{fuzzy}}(B_{m})=\frac{1}{\lvert{\mu_{fuzzy}(B_{m})}\rvert} \sum_{i \in B_{m}} \mu_{fuzzy}(B_{m})\cdot\hat{p}_{i}
\end{equation}

Finally, FCE is calculated as shown in Eq. \ref{fce}. Unlike ECE where the average is taken over the number of samples in $B_{m}$ i.e., $n$, we take the average over the total fuzzy membership in $B_{m}$ i.e., $\sum_{m=1}^{M}\mu_{fuzzy}(B_{m})$.

\begin{equation}\label{fce}
FCE=\frac{1}{\sum_{m=1}^{M}\mu_{fuzzy}(B_{m})}\sum_{m=1}^{M} \lvert{\mu(B_{m})}\rvert\cdot\lvert{\operatorname{acc}{_{fuzzy}}}(B_{m})-{\operatorname{conf}{_{fuzzy}}}(B_{m})\rvert
\end{equation}


\section{Experiments}\label{sec4}

To evaluate the impact of fuzzy binning on calibration error estimation, we perform empirical evaluations across different classification settings. We fine-tune large language models for text classification and measure their calibration performance. 

\subsection{Experimental Setup}
\noindent \textbf{Datasets} 
We consider three text classification datasets to run our analyses, which vary in terms of class distributions, briefly described below.
\begin{itemize}
    \item \textsc{20 Newsgroups (20NG)}: The \textsc{20 Newsgroups} dataset \cite{mitchell1999twenty} is a collection of newsgroup documents containing approximately $20,000$ documents with an (almost) balanced class distribution across $20$ newsgroups/topics. 
    \item \textsc{AGNews (AGN)}: The AG's news topic classification dataset \cite{Zhang2015CharacterlevelCN} is a collection of approximately $128,000$ news articles, from $4$ sources. This dataset is widely used in clustering, classification and information retrieval.
    \item \textsc{IMDb}: The \textsc{IMDb} Movie reviews dataset \cite{maas-EtAl:2011:ACL-HLT2011} is a collection of $50,000$ movie reviews from the Internet Movie Database (IMDb). Each review is assigned either a positive or negative label, and the data is widely used to train models for binary sentiment classification tasks. 
\end{itemize}

We further simulate varying data resource settings to compare miscalibration across different fine-tuning regimes. This is achieved by using a limited portion of the training data to perform fine-tuning, and has been done in prior works \cite{kim_bag_2023}.

\noindent \textbf{Metrics} 
To evaluate calibration across different fine-tuning setups, we use ECE (refer to Eq. \ref{ece}), FCE (refer to Eq. \ref{fce}), and overconfidence (OF), described below.

\begin{itemize}
    \item Overconfidence (\textsc{OF}):  Overconfidence is the expectation of model prediction probabilities $\hat{p}_{i}$ (confidence scores) over incorrect predictions and is calculated as shown in Eq. \ref{of}.

\begin{equation}\label{of}
    \operatorname{OF}=\frac{1}{\left\vert k \right\vert}\sum_{i \in incorrect} \hat{p}_{i}
    \end{equation}

Here $k$ is the total number of incorrect predictions made by a given model.
\end{itemize}

\noindent \textbf{Fine-tuning Setup} 
We implement text classification using a fine-tuned BERT \cite{devlin_bert_2019}. Since the focus of our work is not to create the most accurate fine-tuned model but to compare the efficacy of ECE and FCE across skewed prediction probabilities, we only fine-tune over one epoch and collect miscalibrated prediction probabilities.

\begin{figure}
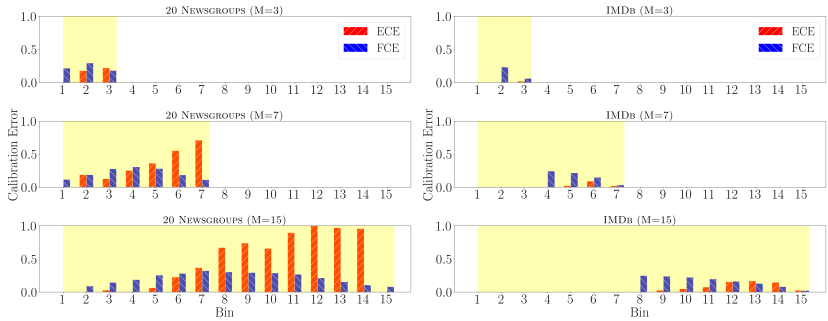

    \centering
    \includegraphics[width=0.45\textwidth]{figures/ce_diffbins_news5k.pdf}
    \includegraphics[width=0.45\textwidth]{figures/ce_diffbins_imdb5k.pdf}
    \caption{Variation in calibration error estimated using ECE and FCE across different bin sizes (top to bottom) and class distributions (left vs right)}
    \label{fig:ce_diffbins}
\end{figure}

\subsection{Results}

\begin{figure}
    \centering
    \includegraphics[width=0.6\textwidth]{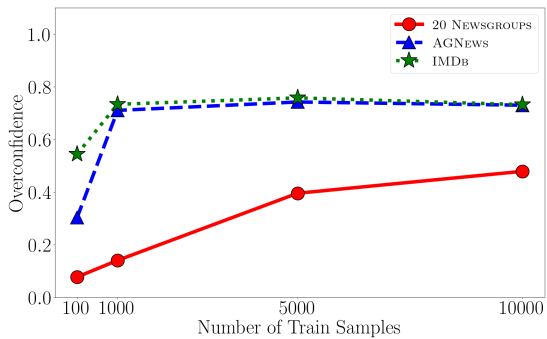}
    \caption{Variation in model overconfidence (OF) across different sample sizes}
    \label{fig:of_samples}
\end{figure}

\textbf{Fuzzy binning in FCE better captures lower probability bins and edge probabilities:} While ECE bins are highly impacted by the leftward skew in prediction probabilities, FCE yields a more uniformly distributed binning scheme. This can be seen in Fig. \ref{fig:ce_diffbins}, where the primary contributors of ECE calculations are the higher probability bins, barely including lower probability bins in calculations. On the other hand, FCE is more uniformly spread across the probability range, better capturing lower probability bins and offering immunity against highly skewed prediction probabilities.

\noindent \textbf{Model overconfidence in multi-class classification settings is low but continuously increasing:} Refer to Fig. \ref{fig:of_samples} to observe the changes in overconfidence in model predictions. Although, a multi-class classification dataset like \textsc{20 Newsgroups} results in lower overconfidence in predictions in limited data regimes, as compared to datasets with fewer classes, this overconfidence increases as the number of samples during fine-tuning increases. On the other hand, datasets with fewer classes i.e., i.e., \textsc{AGNews}  and \textsc{IMDb} output highly overconfident predictions in limited data regimes, but this overconfidence plateaus as one keeps adding more samples.

\begin{table}[]
\centering
    \begin{threeparttable}
        \begin{tabular}{ccccccc}
        \toprule
        \multicolumn{1}{c}{} & \multicolumn{1}{c}{ECE} & \multicolumn{1}{c}{$\Delta_{ECE}$} & \multicolumn{1}{c}{FCE} & \multicolumn{1}{c}{$\Delta_{FCE}$} &  &  \\ \midrule
        Fine-tuning samples        & \multicolumn{4}{c}{\textsc{AGNews}}    &  &  \\ \midrule
        \multicolumn{1}{c}{100}   & 15.41 & \textbf{2.36} & 32.50 & 0.00 &  &  \\
        \multicolumn{1}{c}{1000}  & 3.33  & 0.63 & 11.41 & 0.46 &  &  \\
        \multicolumn{1}{c}{5000}  & 0.71  & 0.41 & 7.77  & 0.71 &  &  \\
        \multicolumn{1}{c}{10000} & 0.80  & 0.78 & 6.86  & 0.66 &  &  \\ \midrule
                                   & \multicolumn{4}{c}{\textsc{IMDb}}      &  &  \\ \midrule
        \multicolumn{1}{c}{100}   & 5.00  & \textbf{1.71} & 22.50 & 0.00 &  &  \\
        \multicolumn{1}{c}{1000}  & 3.42  & 1.51 & 12.01 & 0.24 &  &  \\
        \multicolumn{1}{c}{5000}  & 1.49  & 0.23 & 7.41  & 0.82 &  &  \\
        \multicolumn{1}{c}{10000} & 0.26  & 0.22 & 8.01  & 0.84 &  &  \\ \midrule
                                   & \multicolumn{4}{c}{\textsc{20 Newsgroups}}      &  &  \\ \midrule
        \multicolumn{1}{c}{100}   & 1.31  & 0.20 & 5.90  & 0.00 &  &  \\
        \multicolumn{1}{c}{1000}  & 29.21 & \textbf{4.47} & 38.83 & 0.27 &  &  \\
        \multicolumn{1}{c}{5000}  & 9.99  & 1.54 & 24.05 & 0.11 &  &  \\
        \multicolumn{1}{c}{10000} & 2.28  & 1.30 & 16.18 & 0.39 &  & 
        \\
        \bottomrule
        \end{tabular}%
        \begin{tablenotes}
              \footnotesize
              \item[1] ECE, FCE, $\Delta_{ECE}$ and $\Delta_{FCE}$ values are scaled by a factor of 10. 
              \item
        \end{tablenotes}           
        \caption{Variations in ECE and FCE across different fine-tuning settings. Here, $\Delta$ calculates the average difference in estimated calibration error when binning is performed using fewer bins ($M \in [2..7]$) versus more bins ($M \in [8..15]$). }.
    \end{threeparttable}
\label{tab:ce_table}
\end{table}

\begin{figure}[htbp]%
\centering
\includegraphics[width=0.9\textwidth]{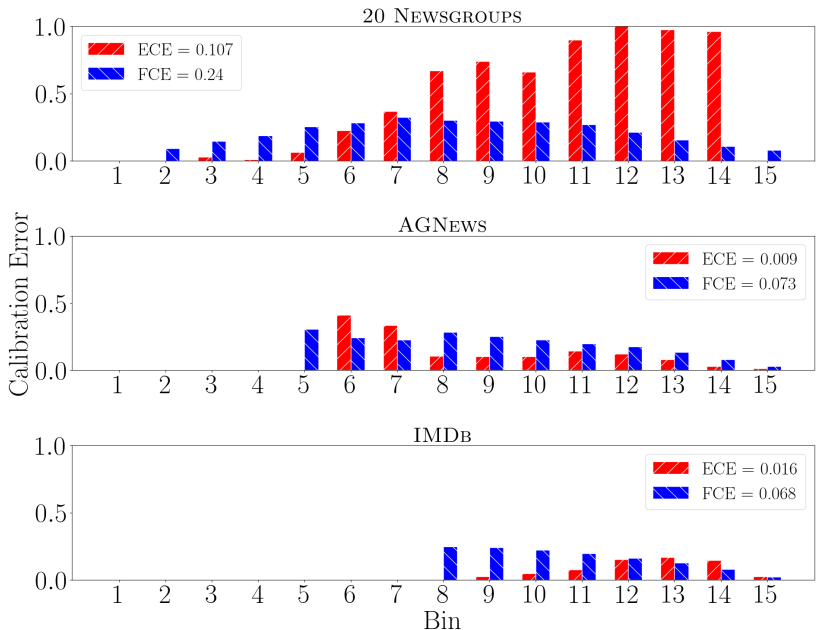}
\caption{Binning of prediction probabilities across $M=15$ bins (model fine-tuned on $n=5000$ samples)}
\label{fig:ce_across_datasets_10k}
\end{figure}

\noindent \textbf{Unlike ECE, FCE is not sensitive to the binning strategy and underlying data used for training:} ECE is a highly sensitive calibration error estimation metric, and is easily influenced by slight changes in data and/or binning strategies. Table \ref{tab:ce_table} shows variations in $\Delta$, which calculates the average difference in estimated calibration error when binning is performed using fewer bins ($M \in [2..7]$) versus more bins ($M \in [8..15]$). While ECE displays larger variations in calibration error estimation due to binning choices, FCE is fairly immune to these choices and shows minimal $\Delta$ in most cases. Further, Fig. \ref{fig:ce_across_datasets_10k} shows that the distribution of ECE across probability bins is highly variable, and usually leftward skewed. On the other hand, FCE bins are more evenly distributed and as shown in Table \ref{tab:ce_table}, output more conservative calibration error estimates.

\section{Conclusion}\label{sec5}
Overconfidence in neural networks lends to the problem of erroneous estimation of calibration error. ECE, a widely adopted metric of measuring calibration error across model decisions has recently come under scrutiny for being biased towards high probability bins. To address this limitation, we propose a new calibration error metric, i.e. Fuzzy Calibration Error (FCE). This metric transforms raw model confidence scores into fuzzy bin memberships, allowing more visibility of lower probability bins within the calibration error calculations. Our results show that FCE offers a tighter estimate of calibration error and the benefits of this metric are more prominent in multi-class classification settings, where skew in model confidence largely affects calibration error estimation using ECE. 
\backmatter





\bmhead{Acknowledgments}

This work was partially supported by the Department of Justice grant \texttt{\#15PJDP-21-GK-03269-MECP}.

\bibliography{sn-bibliography}


\end{document}